\documentclass[conference]{IEEEtran}
\IEEEoverridecommandlockouts
\usepackage{cite}
\usepackage{amsmath,amssymb,amsfonts}
\usepackage{algorithmic}
\usepackage{graphicx}
\usepackage{textcomp}
\usepackage{xcolor}
\usepackage{url}

\def\BibTeX{{\rm B\kern-.05em{\sc i\kern-.025em b}\kern-.08em
    T\kern-.1667em\lower.7ex\hbox{E}\kern-.125emX}}

\usepackage[acronym]{glossaries}

\newacronym{AI}{AI}{Artificial Intelligence}
\newacronym{TACO}{TACO}{Trash Annotations in Context}
\newacronym{CNN}{CNN}{Convolutional Neural Network}
\newacronym{FC}{FC}{Fully Connected}
\newacronym{OvA}{OvA}{One-vs-All}
\newacronym{OvR}{OvR}{One-vs-Rest}
\usepackage{stfloats}

\begin{document}

\title{\textbf{\fontsize{14}{17}\selectfont Efficient Waste Sorting for Circular Economy: A Confidence-guided comparison between One-Vs-All and One-Vs-Rest Classification Strategies with Human-in-the-Loop for Automated Waste Sorting\\}

\thanks{This work has been developed in the project “6RLogistics” (Research Grant Number: 03EI5025B) and is funded by the Federal Ministry for Economic Affairs and Energy (BMWE), Germany.}
}

\author{
    \IEEEauthorblockN{Mohammed Fahad Ali, Dominique Briechle, Marit Briechle-Mathiszig, Tobias Geger and Andreas Rausch}\\
    
    \IEEEauthorblockA{Institute for Software and Systems Engineering\\
                    Clausthal University of Technology\\
                    Germany \\
e-mail: {\{mohammed.fahad.ali}, {dominique.fabio.briechle}, {marit.elke.anke.mathiszig}, \\{thomas.tobias.marcello.geger}, {andreas.rausch}\}@tu-clausthal.de}
}

\maketitle

%%%%%%%%%%%%%%%%%%%%%%%%%%%%%%%%%%%%%%%%%%%%%%%%%%%%%%%%%%%%%%%%%%%%%%%%%%%%%%%%%%%%%%%%%%%%%%%%%%%%%%%%%%%%%%%%%%%%%%%%%%%%%%%%%%%%%%%%%%%%%%%%%%%%%%%%%%%%%%%%%%%%
\begin{abstract}
The complexity of waste disposal regulations across European countries poses significant challenges for the residents and hinders the transition to a Circular Economy. In Germany, the proper sorting and disposal of household waste remains challenging across municipalities. Consequently, substantially reducing incorrectly disposed waste is vital for improving waste management and advancing the Circular Economy. \gls{AI}-based waste sorting solutions can support residents through user-friendly tools, such as mobile applications, that guide proper waste disposal. To be effective in supporting the Circular Economy, however, these solutions must be configurable to reflect the specific waste sorting scheme of individual municipalities in Germany. In the scope of this work, an evaluation and analysis are performed of two prominent classification strategies: \gls{OvA} and \gls{OvR}. The research uses a dataset constructed in alignment with the waste categories and sorting scheme of the city of Goslar in Germany. Moreover, this work aims to extend beyond the overall performance by examining the behavior of \gls{OvA} and \gls{OvR} classification strategies in identifying samples likely to be misclassified. These classification strategies are compared by applying varying confidence thresholds to identify uncertain samples for subsequent human review. This evaluation aims to balance the number of misclassifications against the human effort required for data annotation.
\end{abstract}


%%%%%%%%%%%%%%%%%%%%%%%%%%%%%%%%%%%%%%%%%%%%%%%%%%%%%%%%%%%%%%%%%%%%%%%%%%%%%%%%%%%%%%%%%%%%%%%%%%%%%%%%%%%%%%%%%%%%%%%%%%%%%%%%%%%%%%%%%%%%%%%%%%%%%%%%%%%%%%%%%%%%
\section{Introduction} \label{Introduction}

Climate change is one of the major human-driven challenges of the 21st century \cite{blomeke2020recycling}. In recognition of its importance, the United Nations 2030 agenda includes sustainable consumption and production as one of the 17 Sustainable Development Goals \cite{assembly2015resolution}. In this context, the Circular Economy framework aims for a more sustainable approach by keeping products in use for as long as possible. Unlike its linear counterpart, it focuses on reusing resources in continuous cycles to support sustainable outcomes \cite{lawrenz2021implementing}. In a Circular Economy, effective waste management is vital for maintaining material circulation and reducing resource extraction, and depends partly on proper household waste sorting. Consequently, correctly sorting waste is a structural prerequisite for the Circular Economy, as incorrect sorting contaminates recyclables, raises costs, and often leads to downcycling or disposal.

However, the improper disposal of household waste in Europe is an ongoing issue, causing a massive amount of additional treatment costs and a waste of resources \cite{doi1028650338580}. The overall waste generated by households in the European Union is increasing year by year \cite{eurostat_waste_stats}. In Germany alone, the improper waste disposal in the segment of plastics is up to 30 percent, which corresponds to approximately 750,000 tonnes per year \cite{duale_systeme_2023}. In Europe, and particularly in Germany, the practical implementation of a high-quality recycling workflow is constrained by complex and heterogeneous local regulations, with waste-sorting requirements differing not only between federal states but also between municipalities and even individual districts. For example, in some municipalities, organic waste is sorted together with the residual waste, whereas it is collected separately in other municipalities \cite{volosinova_korinek_2023}.

%Although separated waste collection is already state-of-the-art in European countries, the exact composition of the sorting classes is somewhat unclear.

The member countries of the European Union are trying to tackle both issues with updated regulatory incentives and better coverage of disposal containers \cite{bertelsmann_strategic_planning_guidebook}. However, local regulations and complex institutional settings make such approaches difficult to apply on a large scale. Current citizen information systems, such as print media, therefore, have a limited impact in informing residents about local waste-disposal practices, as they are not easily applied at the point of disposal.

Nevertheless, digital systems can address this lack of applicability by offering easy-to-use tools that increase public knowledge and sorting accuracy, thereby supporting Circular Economy strategies. Consequently, primary APP-based solutions rely on trained \gls{AI} models, most commonly \gls{CNN}s, designed to perform multi-class classification. In this classification setting, deep neural networks are typically configured according to the \gls{OvA} strategy, in which each class in the dataset is assigned a dedicated output unit \cite{PAWARA2020107528}. During inference, the network selects the class with the highest output probability as the predicted label.

Although widely used, a key limitation of the \gls{OvA} classification strategy is that each output unit must learn a complex decision boundary that distinguishes its target class from the union of all remaining classes \cite{PAWARA2020107528}. In addition, modularity and adaptability are other drawbacks with the \gls{OvA} classification strategy. For instance, removing classes for disposal channels requires full retraining or, at a minimum, fine-tuning. Consequently, the need for frequent adaptation to new requirements limits the practical applicability of conventional multi-class classifiers. The technical solution must be customizable to local waste-sorting regulations across municipalities, while remaining usable and beneficial for residents.

An alternative to the \gls{OvA}-based multi-class classification is the \gls{OvR} strategy, which trains separate binary classifiers for each class, thereby enhancing the adaptability and modularity of the overall system. Consequently, retrofitting the classification scheme to comply with new regulations and sorting policies is thereby facilitated, as classes can be selectively dropped or removed without retraining the entire system.

In practical applications, the performance of \gls{AI}-based classification strongly depends on the representativeness of the training data, and misclassifications on previously unseen or under-represented samples are common. Consequently, iteratively augmenting the training set with such unseen or uncertain samples can improve representativeness and thereby reduce misclassification rates. However, this raises the question of how to identify uncertain samples for human review to balance the predictive performance against the data annotation effort. In this context, confidence-guided thresholding is a straightforward strategy, yet its behavior depends on the underlying multi-class decomposition. In the \gls{OvA} classification strategy, the normalized outputs yield probabilities summing to one; thus, low maximum confidence may indicate uncertain samples. In contrast, the \gls{OvR} classification strategy produces independent classifier scores, where conflicting outputs can be used to identify ambiguous samples. Therefore, a systematic comparison of \gls{OvA} and \gls{OvR} in terms of their ability to identify uncertain samples for subsequent human review is crucial to understanding which strategy is more effective for guiding selective annotation in iterative model refinement.

Although promising, it remains unclear whether the performance of these strategies is comparable and whether the advantages of training multiple binary classifiers within the \gls{OvR} classification framework are more efficient than adopting a \gls{OvA} strategy. We are thereby addressing the following two research questions:\\

\begin{enumerate}
    \item How do \gls{OvA} and \gls{OvR} classification strategies compare in terms of overall accuracy and robustness for sorting waste to support Circular Economy goals?\\
    
    \item How does applying a confidence threshold to identify uncertain predictions for human annotation affect the balance between the number of misclassifications and annotation effort, and how does this trade-off differ between \gls{OvA} and \gls{OvR} classification strategies?\\
    
\end{enumerate}

The paper is structured as follows: An overview of the current state-of-the-art is described in Section \ref{Background and State of the Art}. Section \ref{Dataset preparation} outlines the process of data collection to construct the waste management dataset. Subsequently, Section \ref{Framework of Experiments} presents the \gls{AI}-based methodology to compare and evaluate the \gls{OvA} and \gls{OvR} classification strategies, followed by the results and analysis shown in Section \ref{Results and Analysis}. Finally, the findings of both strategies and the potential directions of future research are summarized in Section \ref{Discussion and Future Work}. Lastly, the paper is concluded in Section \ref{Conclusion}.

%%%%%%%%%%%%%%%%%%%%%%%%%%%%%%%%%%%%%%%%%%%%%%%%%%%%%%%%%%%%%%%%%%%%%%%%%%%%%%%%%%%%%%%%%%%%%%%%%%%%%%%%%%%%%%%%%%%%%%%%%%%%%%%%%%%%%%%%%%%%%%%%%%%%%%%%%%%%%%%%%%%%
\section{Background and State of the Art} \label{Background and State of the Art}

The ability to support waste sorting processes is a long-sought-after capability for implementing functional Circular Economy systems, and it has been the subject of several studies. Especially, research in two domains is predominant in the research community: APP-based systems to support citizens in sorting household waste and automated sorting systems for waste recycling processes, both targeting the circularity of products, parts, and resources.

In the case of APP-based platforms, smartphone applications, like \textit{DeepWaste}, support the user to differentiate among three distinct waste types \cite{narayan2021deepwaste}. By taking an image of the waste, the user gets a recommendation on how to sort the garbage into different categories \cite{narayan2021deepwaste}. Moreover, the application \textit{WERTIS-KI}, developed in collaboration with recycling and information technology experts, adopts a comparable approach \cite{wertis_ki_website}. The APP features a chatbot for user interaction that tells users exactly which bin (e.g., paper, plastic, organic waste) to use and shows the walking route to the nearest appropriate trash bin \cite{wertis_ki_website}. Furthermore, the \textit{Junker APP} is another APP that provides helpful information for sorting waste \cite{junker_app}. The APP primarily identifies products by scanning their QR codes (e.g., on bottles, packages, or cans). If no QR code is available, users can take a photo of the item, and the APP uses image recognition to identify the product and suggest proper disposal \cite{junker_app}. Jacobsen et al. \cite{10.11453419249.3420180} propose a different approach with a smart waste bin that uses machine learning models to automate the waste separation process.

When deploying \gls{AI}-based solutions for waste sorting, Aberger et al. \cite{ABERGER2025366} have already proposed an \gls{AI}-powered system to assist manual waste sorting processes. The system relies on \gls{AI}-based classifiers, with MobileNetV2 showing the highest classification accuracy \cite{ABERGER2025366}. In addition, Son and Ahn \cite{SON2025273} examined plastic waste classification using the R-CNN and YOLO v8 model for an automated waste sorting system. The authors \cite{SON2025273} have laid further emphasis on checking for reusability, in case the waste is not labeled as PVC. Similarly, Cheng et al. \cite{su162310155} have proposed an automated waste sorting system using the YOLO v7 model to enable a delta robot to sort different types of beverage cans and bottles.

Regarding different classification strategies, Jang and Kim \cite{DBLP} have examined the applicability of the \gls{OvR} classification strategy for open set recognition as an intelligent self-learning system capable of distinguishing between known and unknown samples. The work introduces a deep neural network architecture that replaces the conventional softmax-based multi-class classifiers with multiple \gls{OvR} classifiers built on top of a \gls{CNN} feature extractor. Each \gls{OvR} classifier uses ReLU activations in its hidden layers and a single sigmoid output for its target class, which enhances learning from non-matching (negative) samples. The experimental results demonstrate more informative hidden representations for the unknown samples than those produced by the standard multi-class classifier. Furthermore, Pawara et al. \cite{PAWARA2020107528} proposed a novel classification strategy, namely One-Vs-One, which was further shown to perform better than the \gls{OvA}-based multi-class classification. Another approach utilizing the \gls{OvR} framework was proposed by Vogiatzis et al. \cite{9651468}, featuring a hierarchical rule framework to support the classification results.

%%%%%%%%%%%%%%%%%%%%%%%%%%%%%%%%%%%%%%%%%%%%%%%%%%%%%%%%%%%%%%%%%%%%%%%%%%%%%%%%%%%%%%%%%%%%%%%%%%%%%%%%%%%%%%%%%%%%%%%%%%%%%%%%%%%%%%%%%%%%%%%%%%%%%%%%%%%%%%%%%%%%
\section{Dataset preparation} \label{Dataset preparation}

\begin{figure}[h]
    \centering
	\includegraphics[width=0.42\textwidth]{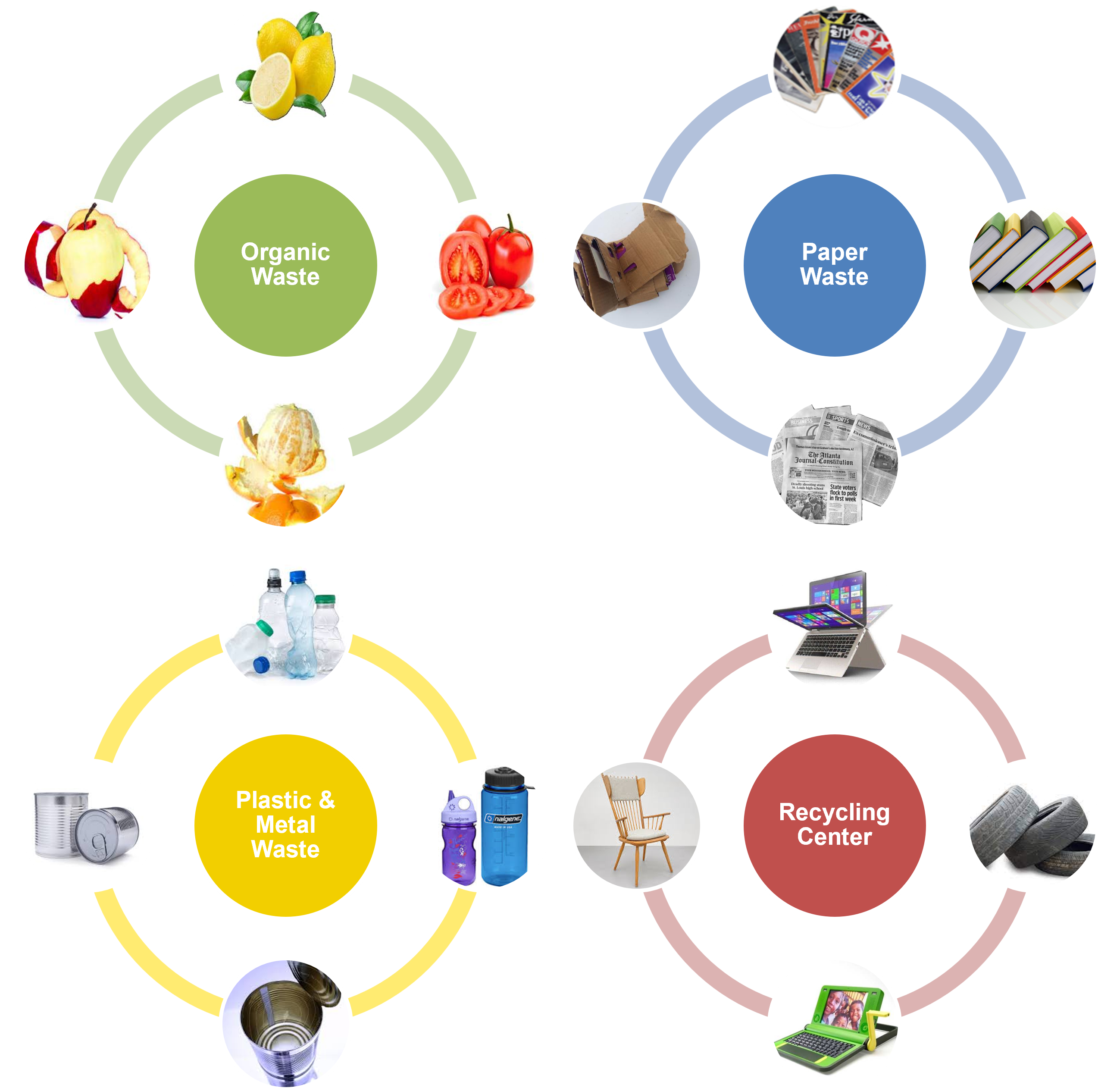}
	\caption{Overview of the four categories in the preliminary version of the waste management dataset.}
	\label{Dataset-V1} 
\end{figure}

This section describes the data collection process and preparation of training, validation, and test sets. The preliminary version of the waste management dataset was constructed by systematically collecting low‑resolution images using techniques such as web crawling. This preliminary version consisted of four distinct categories: organic waste, paper waste, plastic and metal waste, and recycling center. Notably, the defined waste categories follow the sorting scheme specified by the municipal authority of the city of Goslar in Germany \cite{kwb_goslar_abfuhr}. A few examples of the images for each of the four distinct categories are displayed in Figure \ref{Dataset-V1}.

However, the resulting images in the dataset are of heterogeneous quality and limited representativeness, as depicted in Figure \ref{Dataset-V1}. In addition, the displayed waste items are in isolated or unnatural conditions, rather than within the environments where waste is typically encountered. This variability introduces large intra-class differences and visual ambiguities. As a result, it hinders \gls{AI}-based classifiers from learning robust and generalizable features. To mitigate these concerns and enhance the diversity and robustness of the existing dataset, images from two publicly available datasets were considered: \gls{TACO} \cite{TACO-Dataset} and TrashNet \cite{TrashNet-Dataset}.

The \gls{TACO} dataset \cite{TACO-Dataset} is an open-source collection of annotated images of waste. The images are captured in various real-world street environments with considerable variability in illumination, occlusion, and contextual backgrounds. The \gls{TACO} dataset aims to advance research initiatives focusing on effective detection of waste materials \cite{TACO-Dataset}.

On the other hand, the TrashNet dataset \cite{TrashNet-Dataset} is an open-source waste classification dataset featuring objects on a clean background. It was created by Yang et al. \cite{TrashNet-Dataset} from Stanford University in 2016. The TrashNet dataset \cite{TrashNet-Dataset} contains curated images of common waste items in six categories: glass, plastic, paper, metal, cardboard, and trash \cite{TrashNet-Dataset} \cite{Garbage-Classification}. This classification supports effective waste management by enabling accurate identification and sorting of common materials. Consequently, the TrashNet dataset \cite{TrashNet-Dataset} has become highly important in the field of deep learning for garbage classification and is a useful resource for assessing how well neural networks perform on waste sorting tasks \cite{Garbage-Classification}.

\begin{figure}[h]
    \centering
	\includegraphics[width=0.42\textwidth]{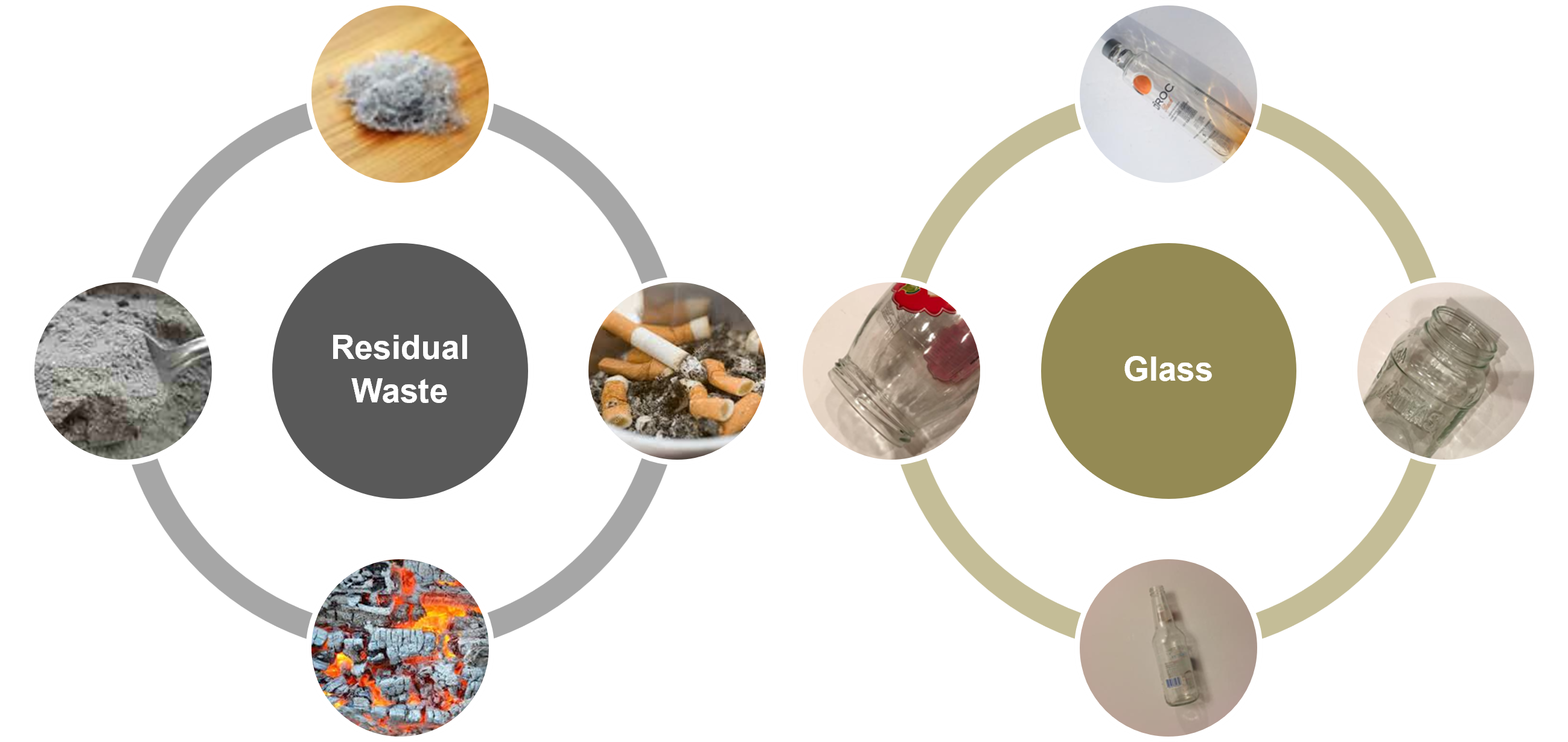}
	\caption{Overview of the extended categories in the waste management dataset.}
	\label{Dataset-V2} 
\end{figure}

In this work, a mapping was defined to align the images from \gls{TACO} and TRASHNET datasets to the existing scheme. The mapping process systematically linked relevant images from these datasets \cite{TACO-Dataset} \cite{TrashNet-Dataset} to the most semantically appropriate category within the established four-class framework. Moreover, to include the unclassified waste items, the existing four-class framework has been expanded with two additional categories: glass and residual waste, as shown in Figure \ref{Dataset-V2}.

Consequently, the dataset extends to a six-class collection with more images and improved representativeness. In addition, incorporating new waste items provides a structured basis for robust waste categorization. Table \ref{tab_dataset_distribution} presents the final waste categories along with the number of images for each category.

\begin{table}[h]
    \centering
    \caption{DISTRIBUTION OF WASTE CATEGORIES ALONG WITH THE RESPECTIVE NUMBER OF IMAGES}
    \begin{tabular}{|c|c|}
        \hline
        \textbf{Waste Categories} & \textbf{Number of Images}\\ 
        
        \hline
        Organic Waste                      & 4634 \\ 
        \hline
        Paper Waste                        & 2591 \\ 
        \hline
        Plastic and Metal Waste            & 3494 \\
        \hline
        Recycling Center                   & 9193 \\ 
        \hline
        Glass                              & 1428 \\ 
        \hline
        Residual Waste                     & 1175 \\ 
        \hline
       
    \end{tabular}
    %\vspace{0.2cm}
    \label{tab_dataset_distribution}
\end{table}

Lastly, a stratified split is performed to ensure proportional representation. This results in 18,515 images for training, 2,000 images for validation, and 2,000 images for testing.

%%%%%%%%%%%%%%%%%%%%%%%%%%%%%%%%%%%%%%%%%%%%%%%%%%%%%%%%%%%%%%%%%%%%%%%%%%%%%%%%%%%%%%%%%%%%%%%%%%%%%%%%%%%%%%%%%%%%%%%%%%%%%%%%%%%%%%%%%%%%%%%%%%%%%%%%%%%%%%%%%%%%
\section{Framework of Experiments} \label{Framework of Experiments}

The designed experimental framework utilizes the constructed waste management dataset to evaluate and compare the \gls{OvA} and \gls{OvR} classification strategies. The dataset consists of six categories with predefined stratified splits, as outlined in Section \ref{Dataset preparation}. To ensure fair and comparable performance evaluations, the same training, validation, and test sets are utilized for both classification strategies. In addition, the hyperparameters are tuned on the same validation set for all trained classifiers, providing a consistent basis for assessing the efficacy of both classification strategies.

To evaluate the \gls{OvA} classification strategy, a single multi-class classifier is constructed. The utilized \gls{CNN} architecture is an InceptionV3 network, which is pre-trained on the ImageNet dataset \cite{ImageNet-Dataset}. In addition, a Global Average Pooling Layer, a \gls{FC} Layer, and a Dropout Layer are added as custom layers on top of the pre-trained \gls{CNN}-based architecture. The final \gls{FC} Layer includes six output units, representing the six distinct classes, followed by the softmax activation function to generate a normalized categorical distribution.

\begin{figure*}[!b]
    \centering
	\includegraphics[width=0.98\textwidth]{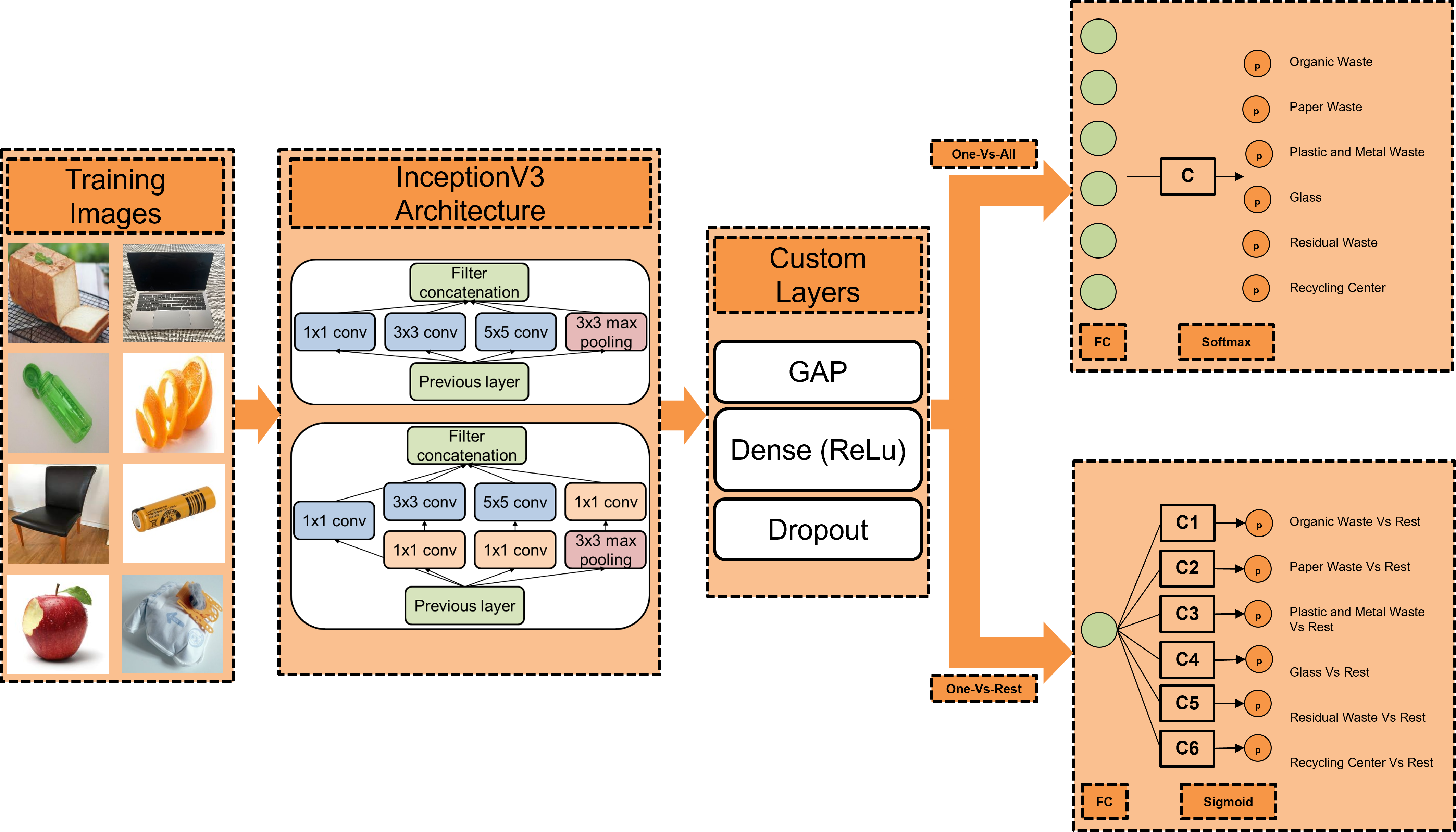}
	\caption{AI-based architecture for evaluating and comparing One-Vs-All (upper branch) with One-Vs-Rest (lower branch) classification strategy.}
	\label{AI-Architecture} 
\end{figure*}

Figure \ref{AI-Architecture} clearly depicts the architecture of \gls{OvA} classification strategy within its upper branch, where a single multi-class classifier is trained. The architecture builds on the pre-trained backbone, followed by a series of custom layers. Subsequently, Figure \ref{AI-Architecture} depicts the final \gls{FC} layer exclusively in the upper branch. In this branch, the final \gls{FC} Layer utilizes a softmax output to implement a standard \gls{OvA}-based multi-class architecture for 6-class classification.

Furthermore, \gls{AI} models are highly configurable through their hyperparameters, which strongly influence the optimal performance \cite{bischl2023hyperparameter}. Consequently, several key parameters are fine-tuned using the designated validation set. This includes choosing the optimizer, setting the learning rate and its scheduler, determining the dropout rate, setting the regularization rate, and selecting the batch size. The fine-tuning process also includes freezing some pre-trained layers and training the remaining ones along with the added custom layers.

On the contrary, to evaluate the \gls{OvR} strategy alongside the multi-class classification, six distinct binary classifiers have been trained, each corresponding to the specific target class. For each binary classifier, positive samples are all training samples from its target class. In contrast, the negative samples encompass all the training samples from the remaining five classes. In addition, each binary classifier utilizes the same pre-trained InceptionV3 backbone along with a consistent set of custom layers. Lastly, the output layer uses a single sigmoid neuron, giving the posterior probability of the target class against all other classes.

Notably, each binary classifier uses the same set of hyperparameters as the multi-class classifier, with values optimized independently on the same validation set. Figure \ref{AI-Architecture} presents the architecture used for all six binary classifiers. The pre-trained backbone is followed by added custom layers, and the final \gls{FC} Layer in the lower branch. In this branch, six independent classifiers, \(C_1, C_2, \dots, C_6\), each with a final sigmoid-activated \gls{FC} Layer, represent the \gls{OvR} classification strategy.

%%%%%%%%%%%%%%%%%%%%%%%%%%%%%%%%%%%%%%%%%%%%%%%%%%%%%%%%%%%%%%%%%%%%%%%%%%%%%%%%%%%%%%%%%%%%%%%%%%%%%%%%%%%%%%%%%%%%%%%%%%%%%%%%%%%%%%%%%%%%%%%%%%%%%%%%%%%%%%%%%%%%
\section{Results and Analysis} \label{Results and Analysis}

This section presents the final evaluation and comparison of the \gls{OvA} and \gls{OvR} classification strategies, based on the classification report and confusion matrix. All the reported performance metrics are derived from the same test set to ensure a fair comparison between the two strategies. In addition, this section analyzes the samples misclassified by both the classification strategies, characterizing their distribution and underlying patterns. Subsequently, a confidence-guided threshold analysis is performed to systematically address the second research question formulated in Section \ref{Introduction}.

%%%%%%%%%%%%%%%%%%%%%%%%%%%%%%%%%%%%%%%%%%%%%%%%%%%%%%%%%%%%%%%%%%%%%%%%%%%%%%%%%%%%%%%%%%%%%%%%%%%%%%%%%
\subsection{One-Vs-All Classification Results} 

\begin{figure}[h]
    \centering
	\includegraphics[width=0.48\textwidth]{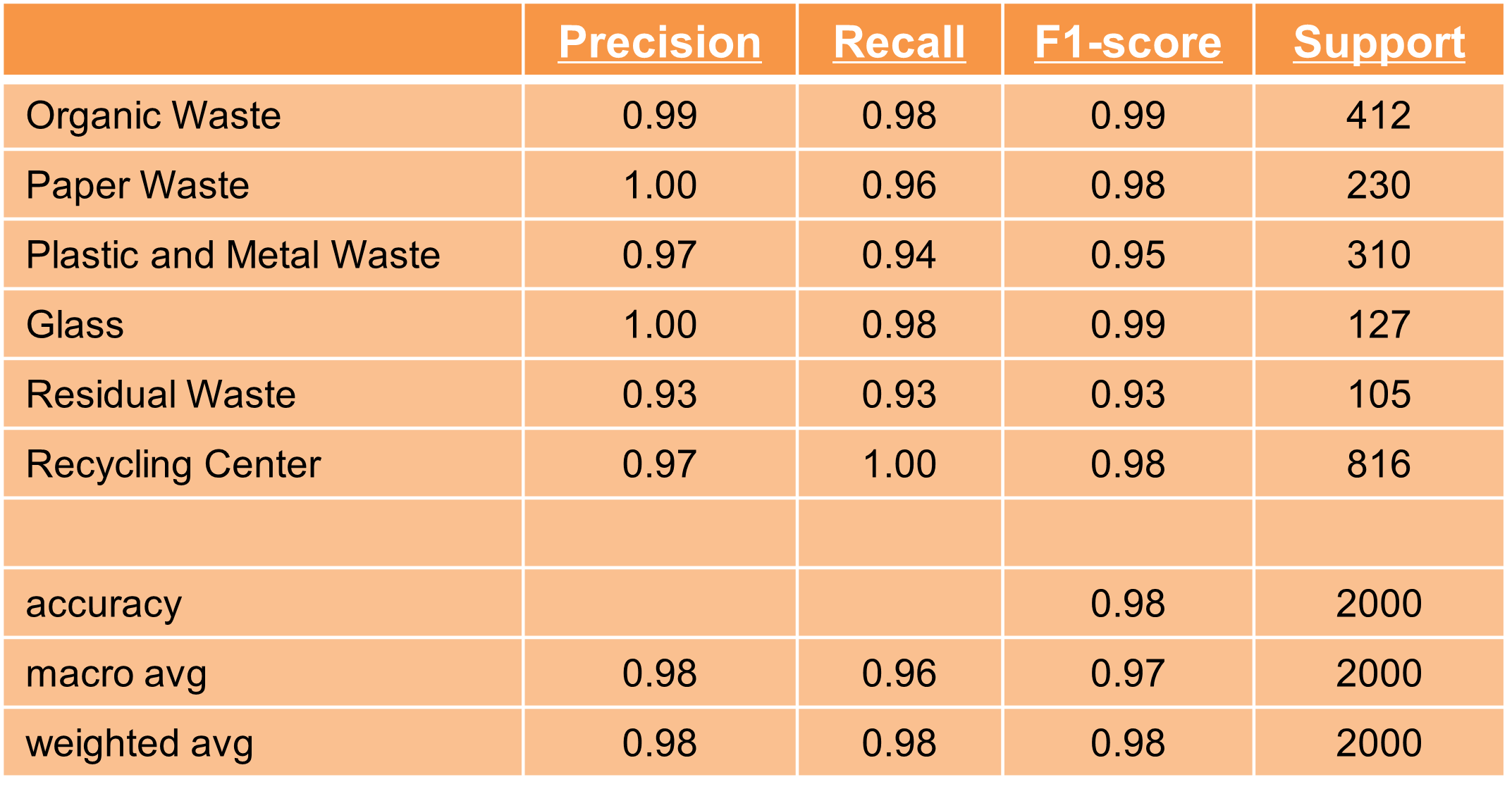}
	\caption{Classification report summarizing the performance of the One-Vs-All classification strategy.}
	\label{CL-OvA} 
\end{figure}

The comprehensive assessment of the predictive performance of the multi-class classifier for sorting waste into six classes is presented in Figure \ref{CL-OvA}. This table presents the key performance metrics: precision, recall, F1 Score, and accuracy.

\begin{figure}[h]
    \centering
	\includegraphics[width=0.42\textwidth]{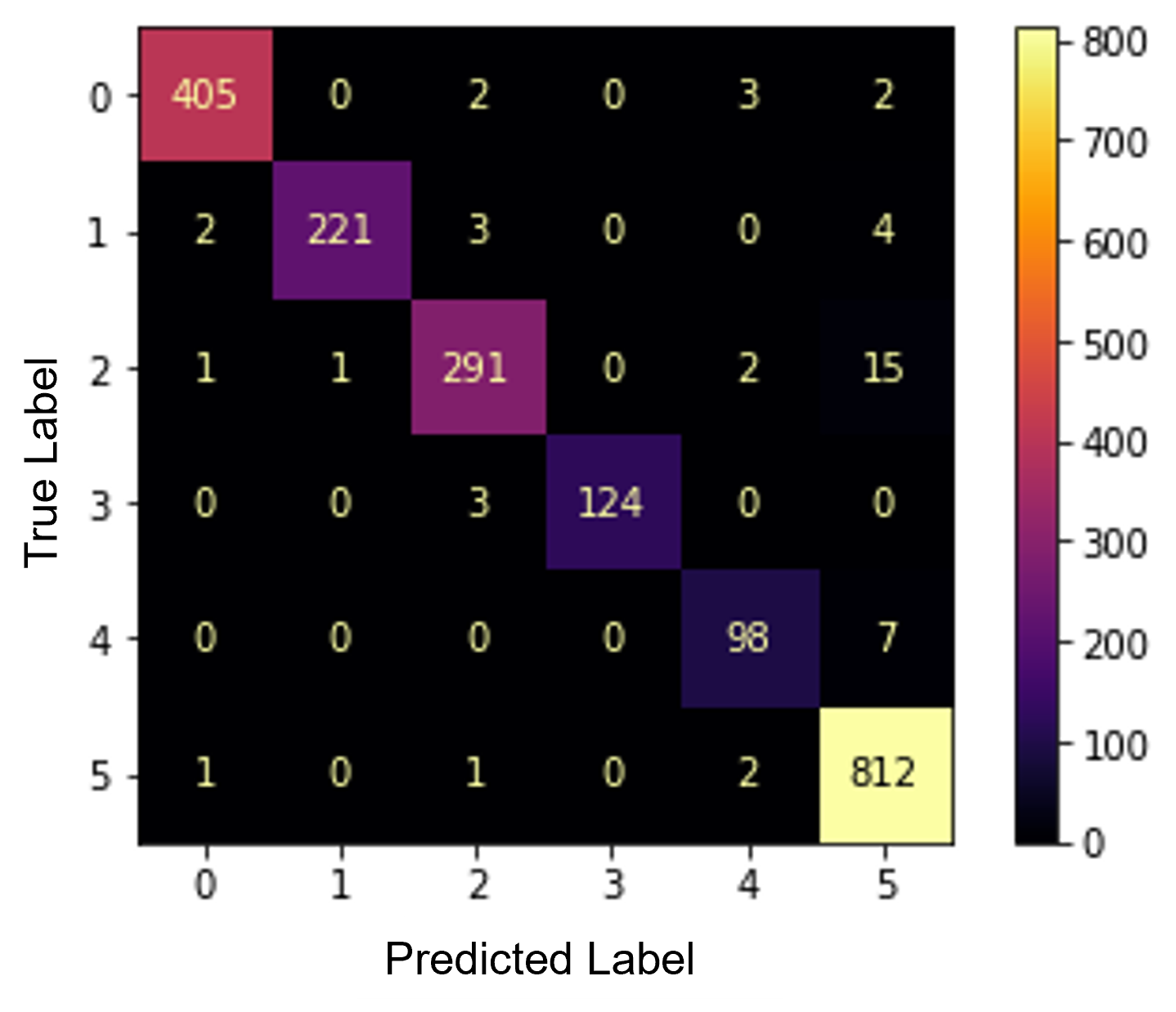}
	\caption{Confusion matrix representing the number of misclassifications of the One-Vs-All classification strategy.}
	\label{CM-OvA} 
\end{figure}

Finally, the confusion matrix is an essential metric for analyzing the instances of misclassifications. The confusion matrix for the evaluation of \gls{OvA} classification strategy is illustrated in Figure \ref{CM-OvA}.

%%%%%%%%%%%%%%%%%%%%%%%%%%%%%%%%%%%%%%%%%%%%%%%%%%%%%%%%%%%%%%%%%%%%%%%%%%%%%%%%%%%%%%%%%%%%%%%%%%%%%%%%%
\subsection{One-Vs-Rest Classification Results}

%As already outlined in Section \ref{Framework of Experiments}, \gls{OvR} classification strategy includes training six distinct binary classifiers, each corresponding to a specific class. In addition, each binary classifier is optimized independently for an identical set of hyperparameters, utilizing the constructed validation set.

\begin{figure}[h]
    \centering
	\includegraphics[width=0.48\textwidth]{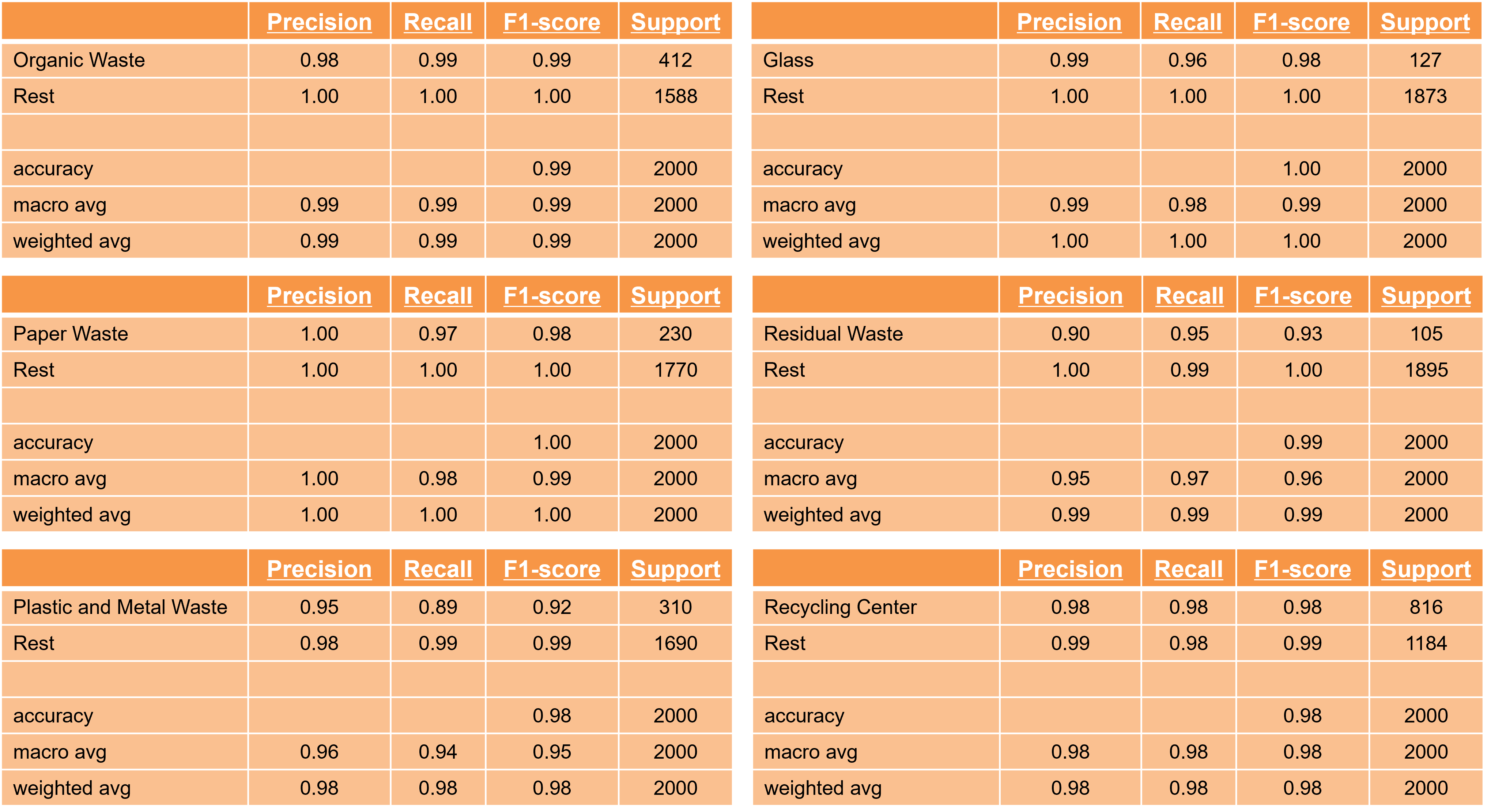}
	\caption{Classification report for each binary classifier within the One-Vs-Rest classification strategy.}
	\label{CL-OvR} 
\end{figure}

The performance of \gls{OvR} classification strategy is illustrated in Figure \ref{CL-OvR}, presenting the classification reports of six independent binary classifiers. The reports summarize the predictive performance for each target class, providing an assessment of the classifier's behavior within the \gls{OvR} framework.

\begin{figure}[h]
    \centering
	\includegraphics[width=0.42\textwidth]{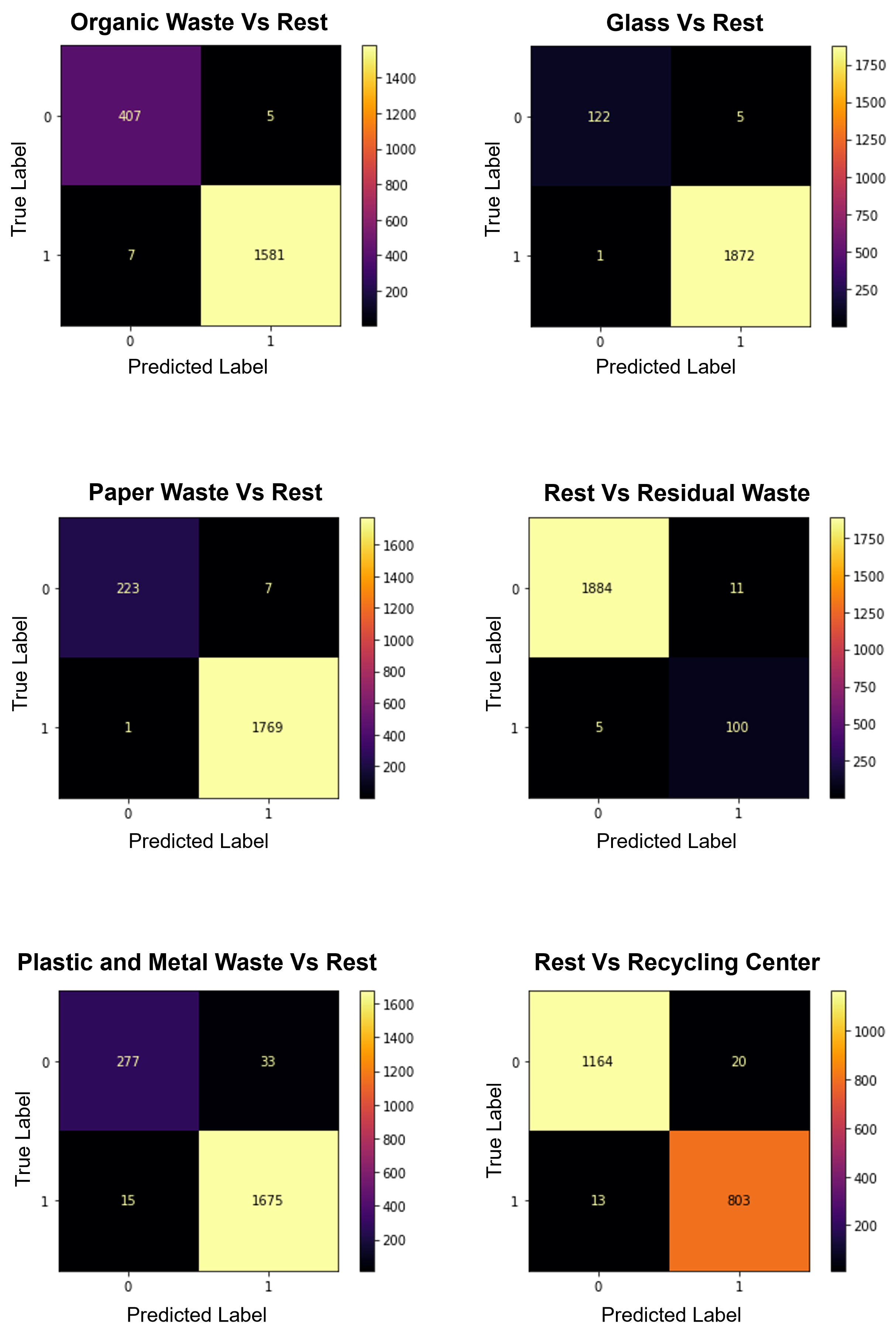}
	\caption{Confusion matrix for each binary classifier within the One-Vs-Rest classification strategy.}
	\label{CM-OvR} 
\end{figure}

Moreover, Figure \ref{CM-OvR} displays confusion matrices for the six binary classifiers within the \gls{OvR} framework. The visualization shows the number of misclassifications for each target class.

%%%%%%%%%%%%%%%%%%%%%%%%%%%%%%%%%%%%%%%%%%%%%%%%%%%%%%%%%%%%%%%%%%%%%%%%%%%%%%%%%%%%%%%%%%%%%%%%%%%%%%%%%
\subsection{Analyzing the misclassified samples}

\begin{figure}[h]
    \centering
	\includegraphics[width=0.44\textwidth]{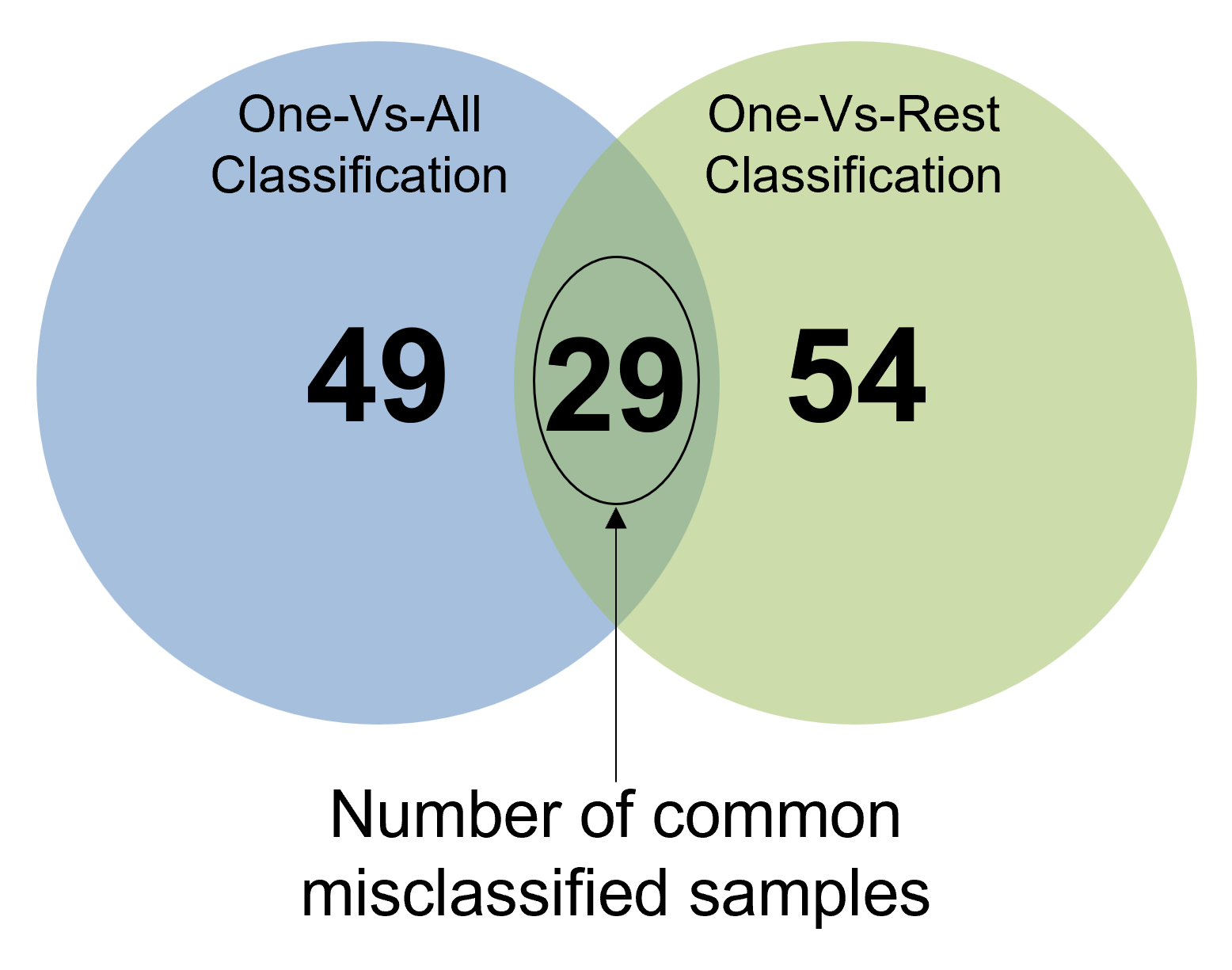}
	\caption{Overview of the number of misclassified samples within the One-Vs-All and One-Vs-Rest classification framework.}
	\label{Number-misclassified} 
\end{figure}

In the performed experiments, the \gls{OvA} classification strategy misclassified 49 images, while the \gls{OvR} classification strategy misclassified 54 images. In addition, among the misclassifications, 29 images were commonly misclassified by both approaches, as illustrated in Figure \ref{Number-misclassified}. The remaining misclassifications, i.e., 20 unique to the \gls{OvA} strategy and 25 unique to the \gls{OvR} strategy, indicate that these approaches exhibit distinct error patterns. Therefore, future work should explore combining both classification strategies in an ensemble framework to improve generalization and robustness.

Furthermore, this analysis goes beyond the overall performance by analyzing the behavior of \gls{OvR} and \gls{OvA} strategies across varying confidence thresholds. This is because some predictions are highly confident and usually correct, while others indicate high uncertainty. Consequently, by partitioning the test set samples into four different probability-based groups, two fundamental questions can be addressed: When the classifier shows high confidence, how reliable is it? In addition, when the classifier generates incorrect predictions, what do its probability estimates indicate about those samples? This advances scientific understanding of both classification strategies and supports practical decisions about the reliability of their predictions.\\

The following are the distinct groups into which the predicted class probabilities for each test sample are categorized:

\begin{itemize}

    \item Group 1 (High-Confidence Samples):
    
    \begin{itemize}
        \item Description: Samples with strong confidence for a single class
        \item Criterion: Exactly one class with prob $\geq$ 0.95; all other classes with prob $\leq$ 0.05
        \item Interpretation: The classifier shows high confidence in the prediction for a single class. However, any misclassification signifies overconfidence, and such samples are "lost" due to high predicted probability.
    \end{itemize}

    \item Group 2 (Single-Vote Samples):

    \begin{itemize}
        \item Description: Samples with a moderate level of confidence for a single class
        \item Criterion: Exactly one class with $0.50 < prob < 0.95$; all other classes with prob $\leq$ 0.50
        \item Interpretation: The classifier shows a moderate level of confidence in the prediction for one of the classes. In this group, wrong predictions represent misclassifications, where the classifier tends to vote for a class with noticeable uncertainty.\\
    \end{itemize}

    \item Group 3 (Multi-Vote Samples): 

    \begin{itemize}
        \item Description: Samples with a moderate level of confidence for multiple classes
        \item Criterion: More than one class with $0.50 < prob < 0.95$; all other classes with prob $\leq$ 0.50
        \item Interpretation: A predicted probability of more than 0.5 is assigned to multiple classes. In the \gls{OvA} strategy, class probabilities sum to 1, so at most one class can exceed 0.5. Therefore, multi-vote samples occur only within the \gls{OvR} classification framework. In the case of the \gls{OvR}, it indicates that the different binary classifiers compete to distinguish certain classes.\\
    \end{itemize}

    \item Group 4 (No-Vote Samples):

    \begin{itemize}
        \item Description: Samples with no dominant class
        \item Criterion: All classes with prob $\leq$ 0.50
        \item Interpretation: The classifier was unable to predict any distinct class with a probability greater than 0.5, indicating a state of high uncertainty, as all probability values remain low.\\
    \end{itemize}
    
\end{itemize}

\begin{figure}[h]
    \centering
	\includegraphics[width=0.48\textwidth]{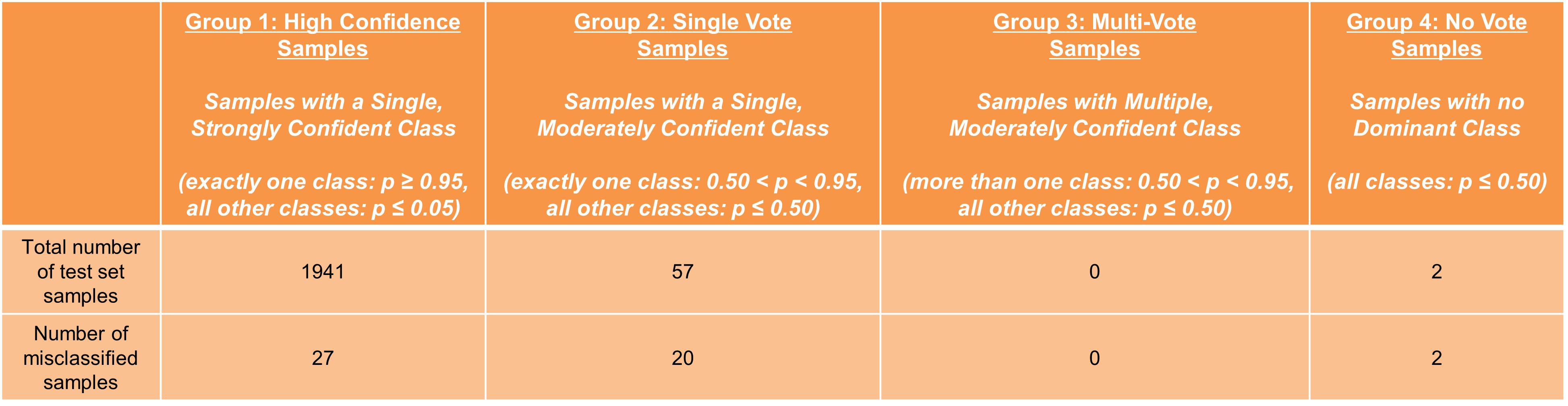}
	\caption{Analyzing the misclassified samples in the specified groups in the One-Vs-All classification strategy.}
	\label{Ergebnisse-OvA} 
\end{figure}

The categorization of the test samples into one of the four distinct groups for the \gls{OvA} classification strategy is presented in Figures \ref{Ergebnisse-OvA}. The results show that there were 27 incorrectly predicted samples out of a total of 1941 samples in Group 1. Consequently, they are difficult to detect using uncertainty heuristics and effectively become "lost" for human correction. Next, Group 2 has a notable 20 misclassifications out of 57 samples. Further, Group 3 has no samples and, as expected, does not contribute to any misclassifications. Finally, Group 4 consists entirely of errors, as both of its 2 samples are misclassified. An expert can review and correct these predictions; however, the impact is expected to be minimal.

\begin{figure}[h]
    \centering
	\includegraphics[width=0.48\textwidth]{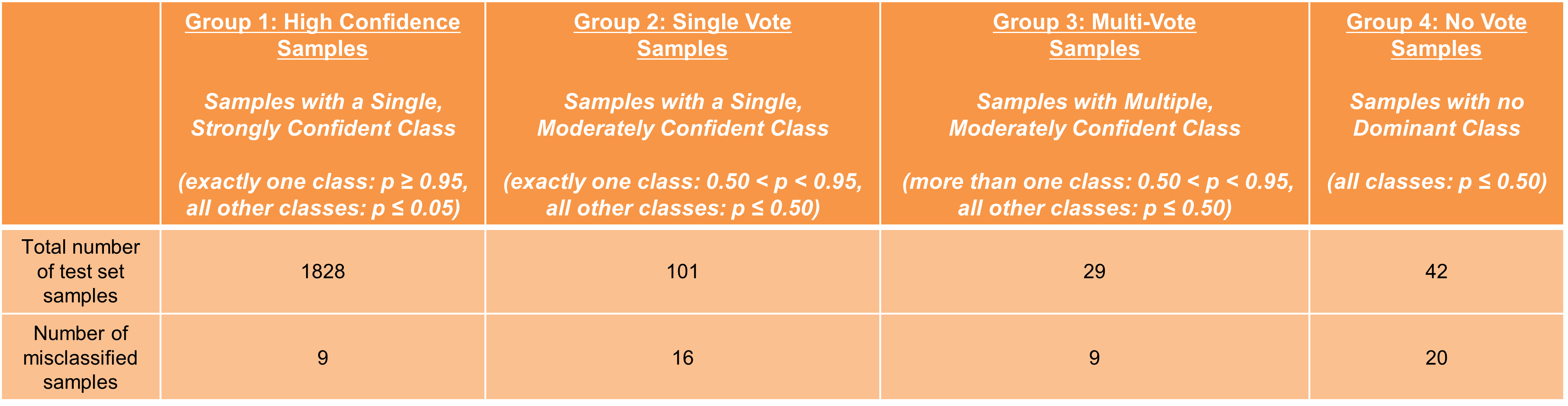}
	\caption{Analyzing the misclassified samples in the specified groups in the One-Vs-Rest classification strategy.}
	\label{Ergebnisse-OvR} 
\end{figure}

Figure \ref{Ergebnisse-OvR} presents the results for the \gls{OvR} classification strategy, with Group 1 including 9 high-confidence misclassifications out of 1828 samples, indicating errors that are difficult to flag and likely to remain uncorrected. Next, Group 2 has 16 misclassified samples out of 101 instances. Further, Group 3 identifies 9 misclassifications out of 29 samples, highlighting it as a priority group for human review. Notably, Group 4 comprises 20 misclassifications out of a total of 42, thereby, small with a high concentration of errors. Consequently, a human annotator can review all its samples, and flagging this group for expert review will reliably reduce misclassifications.

When focusing on Group 4, the \gls{OvA} strategy captures only 2 out of 2000 samples, and both of which are misclassified. This means that 100 percent of the misclassifications can be reduced. However, given the very small sample size, the overall impact remains negligible. In contrast, focusing Group 4 in the case of the \gls{OvR} strategy identifies over 35 percent of all misclassifications while requiring human annotation for less than 2.5 percent of the data.

Moreover, when focusing on Groups 3 and 4, the \gls{OvR} classification strategy requires annotating less than 5 percent of the data, while effectively identifying more than 50 percent of the misclassifications. This indicates that a significant portion of errors originates from a subset of low-confidence predictions. In case of the \gls{OvA} classification strategy, Group 3 has no samples, so the result is equivalent to targeting only Group 4.

%%%%%%%%%%%%%%%%%%%%%%%%%%%%%%%%%%%%%%%%%%%%%%%%%%%%%%%%%%%%%%%%%%%%%%%%%%%%%%%%%%%%%%%%%%%%%%%%%%%%%%%%%%%%%%%%%%%%%%%%%%%%%%%%%%%%%%%%%%%%%%%%%%%%%%%%%%%%%%%%%%%%
\section{Discussion and Future Work} \label{Discussion and Future Work}

%This section can be examined from two complementary perspectives: a technical standpoint and a business-oriented perspective. From a technical point of view,

The conducted experiments demonstrate that the \gls{OvR} strategy misclassified 5 images more than the \gls{OvA} strategy. Thus, the \gls{OvA} strategy performed marginally better than the \gls{OvR} strategy. This outcome emphasizes the effectiveness of the \gls{OvA} strategy in multi-class classification, thereby addressing the first research question.

Furthermore, the \gls{OvR} strategy has proven to be highly effective in selecting uncertain samples for human annotation. The goal is to balance the number of misclassifications with the overall annotation effort. In the case of the \gls{OvR} strategy, when less than 10 percent of the samples are reviewed and corrected by a human expert, focusing specifically on Groups 2, 3, and 4, more than 80 percent of the misclassified samples (45 out of 54) can be captured. Moreover, focusing only on Groups 3 and 4 allows the human expert to annotate fewer than 5 percent of the samples while still capturing more than 50 percent of the misclassified samples (29 out of 54). Lastly, restricting human annotation to Group 4 alone identifies over 35 percent of all misclassifications while necessitating annotation for fewer than 2.5 percent of the data. These outcomes provide evidence in support of the second research question.

Conversely, in the case of the \gls{OvA} classification strategy, targeting Groups 3 and 4 results in only about 4 percent of misclassified samples being captured (2 out of 49). In addition, trusting the samples from Group 1 results in 27 misclassified samples being lost, whereas the \gls{OvR} strategy misses only 9 such samples.

However, the \gls{OvR} strategy is computationally intensive because it necessitates multiple binary classifiers. Nonetheless, the \gls{OvR} classification strategy is more modular in managing individual classes. Another notable finding is that only 29 samples were commonly misclassified by both strategies, indicating that future work on a unified architecture combining both strategies could be beneficial. Another potential direction for future work is to use an adaptive threshold by considering it as a hyperparameter to be optimized. To support the circular economy, future work also focuses on applying the experimental framework to the recycling of electronic circuit boards. This includes constructing an appropriate dataset and sorting the circuit boards into different categories.

To enable the Circular Economy, it is crucial to understand and strategically respond to a decentralized waste management system (like in Germany). Waste management is organized at the local level, meaning that rules can differ from one municipality (i.e., Landkreis) to another. Although national regulations provide a general framework, each municipality establishes its own rules and disposal guidelines. For instance, municipality A may accept "pizza boxes" as paper waste, while municipality B requires them to go in the residual waste. Therefore, to support Circular Economy strategies, residents must follow local guidelines. However, long‑term progress requires greater harmonization of municipal regulations and infrastructure. Consequently, future initiatives could aim to align these regulations within a unified framework to support reusing resources, improve material recovery, and enable a scalable Circular Economy.

%%%%%%%%%%%%%%%%%%%%%%%%%%%%%%%%%%%%%%%%%%%%%%%%%%%%%%%%%%%%%%%%%%%%%%%%%%%%%%%%%%%%%%%%%%%%%%%%%%%%%%%%%%%%%%%%%%%%%%%%%%%%%%%%%%%%%%%%%%%%%%%%%%%%%%%%%%%%%%%%%%%%
\section{Conclusion} \label{Conclusion}

The effectiveness of two classification strategies for waste sorting was evaluated in relation to the formulated research questions. The \gls{OvA} strategy uses a single multi-class classifier to assign probabilities to all classes. Whereas, the \gls{OvR} strategy requires multiple binary classifiers, each separating one target class from all others. The \gls{OvA} strategy achieved marginally higher overall accuracy. However, the \gls{OvR} strategy is particularly effective at identifying uncertain samples and offers modularity that can be tailored to adapt to specific municipal regulations. Although this advantage may come with higher computational costs, a broader evaluation of both strategies is needed to support reconfigurable systems for sorting waste aligned with Circular Economy principles.

%%%%%%%%%%%%%%%%%%%%%%%%%%%%%%%%%%%%%%%%%%%%%%%%%%%%%%%%%%%%%%%%%%%%%%%%%%%%%%%%%%%%%%%%%%%%%%%%%%%%%%%%%%%%%%%%%%%%%%%%%%%%%%%%%%%%%%%%%%%%%%%%%%%%%%%%%%%%%%%%%%%%
\section*{Acknowledgment}
This work was conducted in collaboration with the following partners: Eigenbetrieb Kreiswirtschaftsbetriebe (KWB) Goslar and ceconsoft GmbH.

%%%%%%%%%%%%%%%%%%%%%%%%%%%%%%%%%%%%%%%%%%%%%%%%%%%%%%%%%%%%%%%%%%%%%%%%%%%%%%%%%%%%%%%%%%%%%%%%%%%%%%%%%%%%%%%%%%%%%%%%%%%%%%%%%%%%%%%%%%%%%%%%%%%%%%%%%%%%%%%%%%%%

\end{document}